\documentclass{article}

\usepackage{PRIMEarxiv}              

\usepackage[utf8]{inputenc} 
\usepackage[T1]{fontenc}    
\usepackage{hyperref}       
\usepackage{url}            
\usepackage{booktabs}       
\usepackage{amsfonts}       
\usepackage{nicefrac}       
\usepackage{microtype}      
\usepackage{lipsum}
\usepackage{fancyhdr}       
\usepackage{graphicx}       
\graphicspath{{media/}}     
\usepackage{xspace}[2006/05/08]
\newcommand{\petah}[0]{PETAH\xspace}

\newcommand{\eg}{\textit{e.}\,\textit{g.}\ }
\usepackage{amssymb}
\usepackage{pifont}
\usepackage{array, makecell, multirow, graphicx}
\usepackage{amsmath}
\usepackage{booktabs}
\usepackage{graphicx}
\usepackage{subcaption}

\title{PETAH: Parameter Efficient Task Adaptation for Hybrid Transformers
\thanks{\textit{\underline{Citation}}: 
\textbf{Maximilian Augustin, PETAH: Parameter Efficient Task Adaptation for Hybrid Transformers (CVPRW). Pages 8.  DOI:10.1109/CVPRW67362.2025.00175}} 
}

\author{
  Maximilian Augustin \\
  Tübingen AI Center \\
  University of Tübingen \\
  Tübingen, Germany\\
  \texttt{maximilian.augustin@uni-tuebingen.de} \\
  \And
Syed Shakib Sarwar\\
Meta Inc.\\
Redmond, WA, USA\\
\texttt{shakib7@meta.com}\\
\And
Mostafa Elhoushi\\
Meta Inc.\\
Ontario, Canada\\
\texttt{m.elhoushi@ieee.org}\\
\And
Yuecheng Li\\
Meta Inc.\\
Pittsburgh, PA, USA\\
\texttt{yuecheng.li@meta.com}\\
\And
Sai Qian Zhang\\
New York University\\
New York, NY, USA\\
\texttt{sai.zhang@nyu.edu}\\
\And
Barbara De Salvo\\
Meta Inc.\\
Burlingame, CA, USA\\
\texttt{barbarads@meta.com}\\
}

\begin{document}
\maketitle

\begin{abstract}
Transformers have revolutionized natural language processing (NLP) and are increasingly influential in computer vision tasks. 
Despite their strong performance and multi-tasking capabilities, transformers' high computational demands limit their applicability in resource-constrained environments, where convolutional or hybrid models (combining convolution and attention layers) often excel, particularly in the sub-100M parameter range.
While parameter-efficient task adaptation techniques have been successful in NLP, they have not been widely adopted for hybrid transformers in vision tasks. 
In this work, we introduce \petah (Parameter Efficient Task Adaptation for Hybrid Transformers), a novel framework for efficiently adapting hybrid transformers to new tasks. 
We further combine \petah with pruning to create high-performing and storage-efficient models suitable for multi-tasking. 
Our extensive evaluations on classification and other vision tasks demonstrate that \petah-adapted hybrid models outperform established task-adaptation techniques for Vision Transformers (ViTs), requiring fewer parameters and achieving greater efficiency on mobile hardware.
\end{abstract}


\section{Introduction}
\label{sec:Introduction}

Transformers~\cite{vaswani2017attention} have achieved remarkable success in both NLP and computer vision, excelling in tasks such as classification~\cite{dosovitskiy2020image, steiner2021train, fang2022eva, fang2023eva}, semantic segmentation~\cite{cheng2021mask2former}, and object detection~\cite{carion2020end, zhu2020deformable, meng2021conditional}. Recent advancements like DINOv2~\cite{oquab2023dinov2} demonstrate that large-scale Vision Transformers (ViTs) can be transferred to multiple downstream tasks without extensive retraining. This approach uses the pre-trained model as a fixed feature extractor, allowing task-specific heads to be trained independently.

However, these capabilities are limited to large ViTs, trained on large-scale datasets and with hundreds of millions of parameters, which are unsuitable for resource-constrained applications. In contrast, hybrid models that combine convolutional and attention layers have shown superior performance in the sub-100M parameter range~\cite{mehta2021mobilevit, li2022efficientformer, li2022rethinking,  mehta2022separable, li2022next, vasu2023fastvit}. Despite their efficiency, parameter-efficient task adaptation techniques, which are even more useful for storage-limited mobile applications, have not been widely applied to these hybrid architectures, creating a significant gap in the current research landscape.

In NLP, parameter-efficient adaptation methods like Adapters~\cite{houlsby2019parameter} and Low-Rank Adaptation (LoRA)~\cite{hu2021lora} have successfully adapted LLMs to new tasks without retraining entire networks, offering significant storage savings and flexibility. Extending these benefits to hybrid transformers in vision tasks could enable a single backbone to serve multiple applications. This is particularly beneficial in mobile settings with stringent storage limitations. This approach would allow hardware manufacturers to deploy a single, optimized network capable of handling multiple tasks, thereby reducing storage requirements and simplifying updates.

\begin{figure}[t]
    \centering
    \includegraphics[width=\textwidth]{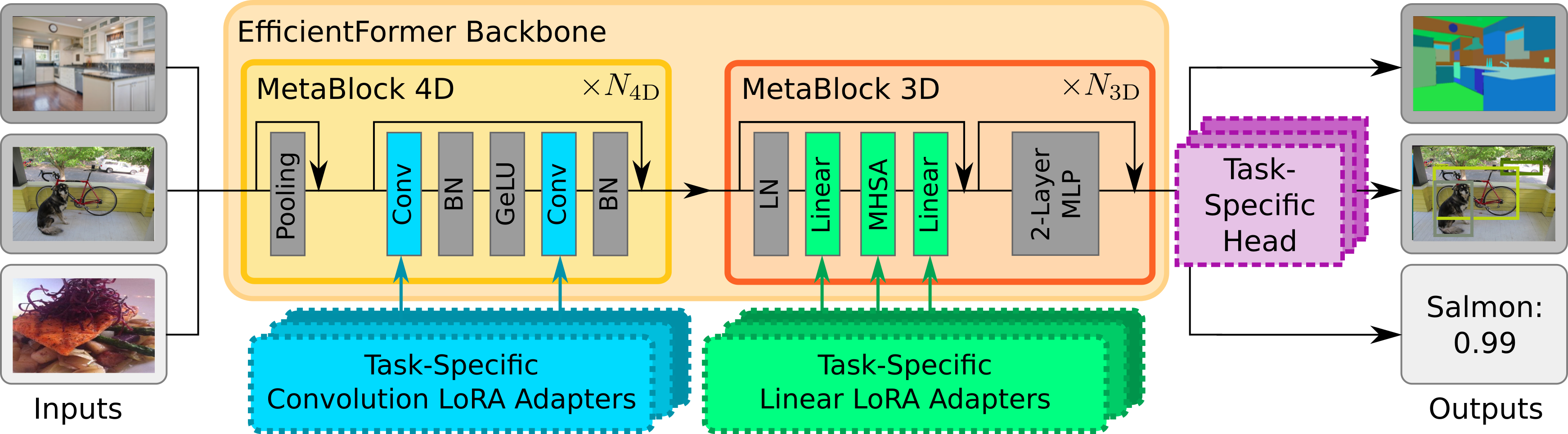}
    \caption{Illustration of our PETAH framework for adapting hybrid transformer models. PETAH employs low-rank adaptation to modify both fully connected and convolutional layers, enabling EfficientFormer to tackle various computer vision tasks (\textit{e.g.}, fine-grained classification, detection, segmentation) efficiently. Task-specific learnable parameters, including LoRA adapters and task heads, are highlighted with dashed borders, while other parameters (\textit{e.g.}, convolution and linear layer weights) remain fixed. Color coding denotes adapted modules, with dark-gray indicating frozen modules like layer-norm (LN) and MLP. }
    \label{fig:teaser}
\end{figure}

In this work, we introduce \petah (\textbf{P}arameter \textbf{E}fficient \textbf{T}ask \textbf{A}daptation for \textbf{H}ybrid Transformers), the first framework to efficiently adapt hybrid transformer architectures to a diverse set of new vision tasks (see Figure \ref{fig:teaser}). Our main contributions are as follows:

\begin{enumerate}
    \item \textbf{Adaptation Framework:} We propose \petah, enabling efficient task adaptation for hybrid transformers by modifying both convolutional and attention layers.
    \item \textbf{Performance Gains:} We demonstrate that \petah-adapted EfficientFormer models outperform standard ViT-B models with various PEFT approaches by nearly 1\% in average accuracy while being over twice as fast on mobile NPUs.
    \item \textbf{Efficiency Through Pruning:} By combining \petah with pruning techniques, we create highly efficient models in the sub-10M parameter range, highlighting the synergy between task adaptation and sparsity. Our pruned EfficientFormer has half as many parameters as a standard ViT-S, achieves faster on-device latency, and has 1.5\% higher average accuracy.
    \item \textbf{Versatility Across Tasks:} We extend our evaluation to dense tasks like semantic segmentation and object detection, showing that \petah can adapt hybrid backbones without inference-time compute overhead, matching the performance of full fine-tuning.
\end{enumerate}

Our extensive evaluations demonstrate that \petah not only improves performance on classification tasks but also extends to dense prediction tasks, making it a versatile and efficient solution for adapting hybrid transformers in resource-constrained settings.

\section{Related work}
\label{sec:Related Work}

\textbf{Vision Architectures:} For nearly a decade after the introduction of AlexNet~\cite{krizhevsky2012imagenet}, computer vision research focused on convolutional networks such as ResNets~\cite{he2016deep}, MobileNets~\cite{sandler2018mobilenetv2, howard2017mobilenets}, and EfficientNets~\cite{tan2019efficientnet}. Transformers~\cite{vaswani2017attention} were initially proposed for handling long text sequences but quickly gained traction in the vision domain with the advent of ViTs~\cite{dosovitskiy2020image}. Due to their architecture, ViTs have fewer inductive biases and thus necessitate the development of new training schemes, including the use of strong augmentations~\cite{steiner2021train}, distillation~\cite{touvron2021training}, and self-supervised learning~\cite{he2022masked, caron2021emerging, fang2022eva, wang2023one}. Further research has focused on refining the original ViT architecture~\cite{touvron2021going, liu2021swin, wang2021pyramid} and modifying attention mechanisms~\cite{yu2022metaformer, liu2021swin, pan2022fast, kitaev2020reformer, chen2021regionvit}.

While ViTs demonstrate great performance at scale, they cannot compete with efficient convolutional architectures in resource-constrained settings when it comes to inference speed and parameter efficiency~\cite{wang2022towards, marin2021token}. To bridge this gap, recent works introduced hybrid transformers~\cite{dai2021coatnet, graham2021levit, mehta2021mobilevit, mehta2022separable, vasu2023fastvit, wu2022tinyvit, li2022rethinking, li2022efficientformer, chen2022mobile, cai2022efficientvit}, which combine convolutional and attention-based layers. Models such as FastViT~\cite{vasu2023fastvit}, EfficientFormer~\cite{li2022efficientformer, li2022rethinking}, and TinyViT~\cite{wu2022tinyvit} have proven to be more parameter and compute-efficient than ViTs, making them the best available models for resource-limited settings.

\textbf{Multi-Tasking and Task Adaptation:} Multi-task learning~\cite{caruana1997multitask, vandenhende2021multi} refers to one model solving multiple tasks, often by sharing a fixed image encoder that branches out into task-specific heads~\cite{chen2018gradnorm, kendall2018multi, sener2018multi}. This requires careful task and gradient weighting during optimization~\cite{kendall2018multi, guo2018dynamic, liu2019end, chen2018gradnorm, zhao2018modulation}. As traditional multi-task learning is often formulated as a multi-objective optimization problem, all tasks have to be known before training. Adding a new task generally requires joint retraining across tasks, which is impractical in many applications where a system-wide vision backbone should be available for various downstream applications.

In NLP, the scale of large language models has made full fine-tuning impractical, leading to the development of parameter-efficient fine-tuning (PEFT) methods~\cite{hu2021lora, houlsby2019parameter, liu2023gpt, he2023parameter, zaken2021bitfit}. While prompt and prefix-based approaches~\cite{li2021prefix, liu2023gpt} can be hard to generalize to vision tasks, adapter-based approaches like LoRA~\cite{hu2021lora} and its variants~\cite{chavan2023one, edalati2022krona, valipour2022dylora} can be adapted to models containing attention layers.

Recent works have applied PEFT to ViTs~\cite{he2023parameter, chavan2023one}, but there is a lack of research on adapting hybrid transformers, which combine both convolutional and attention layers. Our work fills this gap by introducing \petah, a PEFT framework specifically designed for hybrid transformers.

\section{Preliminaries}
\label{sec:Preliminaries}

\subsection{Hybrid architectures and pre-training}\label{sec:hybrid_architectures_and_pre_training}
To fairly evaluate different task-adaptation methods for hybrid transformers and compare them to task-adaptation on vision transformers, we have to select a hybrid architecture and a pre-training framework that we can use to train all models. For most of our experiments, we will use the EfficientFormer (EF) \cite{li2022efficientformer} in both the L3 variant with 31M and the L7 variant with 80M parameters. Unlike some of the later hybrid variants \cite{li2022rethinking, vasu2023fastvit, wu2022tinyvit}, the EF contains transformer blocks that are very similar to that of a standard ViT which makes it easy to adopt NLP task-adaptation techniques and compare to a ViT. In particular, the EF consists of a convolutional stem and three stages of 4D MetaBlocks, that while inspired by the transformer design, use convolutions to operate on a batch of 3D feature maps ($C \times H \times W$ with channels $C$, width $W$ and height $H$). In the final stage, the 3D feature maps are flattened into a 2D sequence of size $( H \cdot W ) \times C$ which is processed by multiple 3D MetaBlocks that contain a standard multi-head attention layer with an additional linear projection and a two-layer MLP. In contrast, the standard ViT only consists of a patchify layer and multiple transformer blocks, each containing an attention module and a two-layer MLP. Thus the fourth stage of the EF model strongly resembles the standard ViT architecture, however, the first three stages are built using faster and more parameter-efficient convolutional layers. While the EF is not the most parameter-efficient hybrid model \cite{li2022rethinking, vasu2023fastvit}, it is much more efficient in terms of number of floating point operations and on-device latency than ViTs.

Next, we have to define the pre-training setup. Although initially most image classifiers were trained on ImageNet-1K,  the vision community has adapted several new pre-training datasets and algorithms over the past years. While pre-training in a self-supervised fashion on massive datasets containing hundreds of millions or even billions \cite{schuhmann2022laion} of images has proven to produce models that can easily adapt to new downstream tasks \cite{oquab2023dinov2, sun2023eva, fang2023eva, wang2023one}, training such models can be prohibitively expensive. Since we are interested in adaptation to a wide variety of downstream tasks, we chose ImageNet-21K pre-training as it strikes a balance between being sufficiently general without being too large.
We use the DeiT III \cite{touvron2022deit3} framework to train several hybrid models and ViT baselines. For all models, we use the exact same training setup and train for 90 epochs using a resolution of $224 \times 224$ and the recommended parameters from \cite{touvron2022deit3}. Since the original fall release of IN21K is no longer available, we follow \cite{ridnik2021imagenet} and use their subset with 10450 classes. By training both the ViT baseline and the hybrid models on the same dataset with the same augmentation and parameters, we can directly compare task-adaptation performance between these two architectures without other factors influencing the analysis. 

The validation results on ImageNet-21K, as well as parameter counts, compute costs, and on-device latency on an iPhone neural processing unit (NPU), are presented in Table \ref{tab:our_backbones}. 
Notably, the EF L7 is a great model to compare to the ViT-B in terms of task adaptation performance since both have a similar parameter count and ImageNet-21K validation accuracy. However, the EF L7 has substantially fewer floating-point operations and is faster than even a smaller ViT-S on a mobile NPU.
\begin{table}
    \centering
    \caption{Comparison of backbones used in subsequent experiments. All models were trained for 90 epochs on ImageNet-21K with identical hyperparameters. NPU latency results are sourced from \cite{li2022efficientformer, li2022next}.}     
    \begin{tabular}{l|c|c|c|c}
        \hline
        \textbf{Model} & \textbf{IN21K acc.} & \textbf{\#Params} & \textbf{Gflops} & \textbf{NPU Latency} (ms)  \\
        \hline
        EfficientFormer L7  & 50.43 & 80M & 20.3 & 6.9\\ 
        EfficientFormer L3& 47.72 & 31M & 7.84 & 3.0\\ 
        ViT-B &  49.99 & 85M & 35.13 & 18.2\\ 
        ViT-S &  46.02 & 22M & 9.20 & 9.0\\ 
        \hline
    \end{tabular}
    \label{tab:our_backbones}
\end{table}
\section{Task Adaptation for Hybrid Transformers}
\label{sec:4_Task Adaptation for Hybrid Transformers}

\begin{table}
    \centering
    \caption{Adaptation of an EfficientFormer L7 pre-trained on ImageNet-21K to fine-grained classification tasks using standard PEFT approaches (focusing on fully connected layers in transformer blocks) and our version incorporating convolutional LoRA adaptation (LoRA ATTN + Conv LoRA). For comparison, higher-rank LoRA variants and a LoRA variant adapting both attention and MLP layers are included. Despite increased parameter counts, these variants underperform compared to adding low-rank convolutional adaptation. }
    \label{tab:classification_lora_layers}
    \begin{tabular}{l|ccc|c|c}
        \hline
         \textbf{Type} &  \textbf{Aircraft} & \textbf{DTD} & \textbf{Food} & \textbf{Mean} & \textbf{\#Params}\\
         \hline
         Linear Probing & 52.98 & 75.60 & 86.65 & 71.74 & 0\\
         \hline
         LoRA ATTN $r=8$ & 69.20 & 76.83 & 89.12 & 78.38 & 0.26M\\
         LoRA ATTN $r=16$ & 70.53 & 76.67 & 89.37 & 78.85 & 0.52M\\
         LoRA ATTN + MLP $r=8$ & 69.20 & 77.27 & 89.31 & 78.59 & 0.75M\\
         LoRA ATTN + MLP $r=16$ & 69.37 & \textbf{77.45} & 89.53 & 78.78  & 1.5M\\
         \hline
         LoRA ATTN $r=8$ &&&&&  \\
         \ + Conv LoRA $r_c=1$ & 75.69 & 75.50 & \textbf{90.77} & 80.65 & 0.35M \\
         \ + Conv LoRA $r_c=2$ & \textbf{75.96} & 77.32 & {90.66} & \textbf{81.31} & 0.45M \\
    \hline
    \end{tabular}
\end{table}

\subsection{Task adaptation methods}
We briefly recap existing methods for transformers that we consider as baselines for our work. For transformers, it is common~\cite{he2023parameter, hu2021lora, touvron2022three} to adapt only the attention modules without adapting the MLP modules. 
Unless otherwise stated, we assume that we want to adapt a linear transformation $f(x) = W_0x + b$, parameterized by the weight matrix $W_0 \in \mathbb{R}^{p \times q}$ and bias vector $b \in \mathbb{R}^{p}$ where $p$ and $q$ are the output and input dimensions.

\subsubsection{LoRA - Low-Rank Adaptation} 
LoRA \cite{hu2021lora} is one of the most popular parameter-efficient methods introduced for the adaptation of large language models. Given a pre-defined rank $r$ the modified forward pass of the linear transformation is defined via the weight update $\Delta W$ which is given as the outer product of two low-rank matrices $A \in \mathbb{R}^{r \times q}$ and $B \in \mathbb{R}^{p \times r}$ as $\Delta W = B A$, thus:

\begin{equation}\label{eq:lora}
        \begin{split}
        W_0 x + \Delta W x + b &= W_0 x + B A x + b\\
        &= (W_0  + B A) x + b
    \end{split}
\end{equation}

Importantly, since the updated weight matrix $W_0  + B A$ can be computed while loading the model weights and adapter parameters, LoRA does not introduce any computation overhead during inference. 

\subsubsection{LoRA for convolutional layers}
Due to their origins in NLP, most PEFT methods target the transformer architecture and its attention layers. It is less commonly known that convolutional layers can also be adapted using such decompositions \cite{lebedev2014speeding, phan2020stable}.  Assume we are given a convolutional layer with kernel size $k$, $p$ output- and $q$ input channels parameterized by the weight tensor $W_\textrm{4D} \in \mathbb{R}^{p \times q \times k \times k}$ and bias $b \in \mathbb{R}^{p}$. We use $\textrm{conv2D}(\cdot,\cdot) $ to denote the function that applies a 2D convolution specified by the kernel in the second argument to the input given as the first argument. 

To apply PEFT methods designed for fully connected layers to convolutional layers, we can flatten the 4D tensor and reshape it to a standard matrix $W_\textrm{2D}$ of size $p \times (q \cdot k^2)$. If we want to apply LoRA with rank $r$ to $W_\textrm{2D}$, the dimensions of the two low rank matrices $B_\textrm{2D}$ and $A_\textrm{2D}$ are $p \times r$ and $r \times (q \cdot k^2)$. The resulting weight update is again given via the standard matrix product: $\Delta W_\textrm{2D} = B_\textrm{2D} A_\textrm{2D}$.
Following \cite{wang2021pufferfish}, we can now reshape $A_\textrm{2D} \in \mathbb{R}^{r \times (q \cdot k^2)}$ back to a 4D tensor $A_\textrm{4D} \in \mathbb{R}^{r \times q \times k \times k}$. Similarly, $B_\textrm{2D} \in \mathbb{R}^{p \times (r \cdot 1^2)}$ can be reshaped to a $1\times1$ convolution kernel: $B_\textrm{4D} \in \mathbb{R}^{p \times r \times 1 \times 1}$. Thus, analogously to standard LoRA \ref{eq:lora}, we have the convolutional LoRA version:

\begin{equation}\label{eq:conv_lora}
   \textrm{conv2D}(x,W_\textrm{4D}) + \textrm{conv2D}\big(\textrm{conv2D}(x,A_\textrm{4D}),B_\textrm{4D}\big) + b.
\end{equation}

After training $A$ and $B$ in their 4D tensor representations, we can reshape them back to their 2D version and calculate $\Delta W_\textrm{2D} = B_\textrm{2D} A_\textrm{2D}$. We can reshape $\Delta W_\textrm{2D}$ to the 4D kernel representation $\Delta W_\textrm{4D}$ and add it back to the original weight $W_\textrm{4D}$ to obtain the final kernel: $W_\textrm{4D} + \Delta W_\textrm{4D}$. A single convolution with this modified kernel is then equivalent to \ref{eq:conv_lora}, thus during inference time, the updated forward pass is given by: 

\begin{equation}\label{eq:conv_lora_inference}
   \textrm{conv2D}(x,W_\textrm{4D} + \Delta W_\textrm{4D}) + b.
\end{equation}

Thus, like standard LoRA, convolutional LoRA does not increase inference time. While there have been works demonstrating the application of LoRA-like methods to convolutional layers \cite{hyeon2021fedpara, yeh2023navigating, zhong2024convolution} for generative models, segmentation, and federated learning, up to our knowledge, there does not exist any work demonstrating the feasibility of using convolutional PEFT for computer vision multi-tasking and/or adapting hybrid transformers. 

\subsubsection{Other PEFT methods} While LoRA remains the most popular PEFT method, many follow-up papers attempted to improve the parameter efficiency and performance, \eg Kronecker Adaptation \cite{edalati2022krona} or LOHA \cite{yeh2023navigating}. As the main goal of our work is to demonstrate that hybrid models profit not only from adapting attention layers but also convolutional layers and can serve as flexible backbones for multiple vision tasks, we consider these approaches orthogonal to our objectives and focus on LoRA-based methods. With the reshaping formulation outlined in the previous paragraph, it is possible to adapt \textit{any} PEFT method based on matrix factorizations to convolutional layers and, while we consider this out-of-scope for this work, it can be a promising direction for further research.

\subsubsection{Attention fine-tuning} Since the MLP layers in a transformer tend to be parameter-heavy, \cite{touvron2022three} proposed to only finetune the attention layers. They demonstrate that this approach can save memory and compute during fine-tuning while maintaining the accuracy of full fine-tuning. While attention tuning is still parameter-intensive—for example, in a standard ViT-B, the attention layers contain about 28M parameters—the authors of \cite{touvron2022three} argue that, in terms of performance, for ViTs, \textit{``fine-tuning attention is all you need"}.

\begin{figure}
    \centering
     \centering
     \begin{subfigure}{0.49\textwidth}
         \centering
         \includegraphics[width=\textwidth]{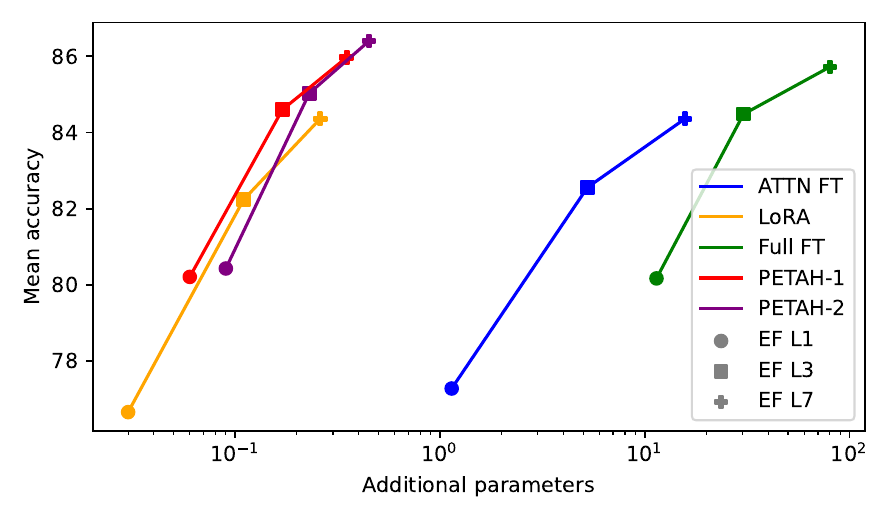}
     \end{subfigure}
      \centering
     \begin{subfigure}{0.49\textwidth}
         \centering
         \includegraphics[width=\textwidth]{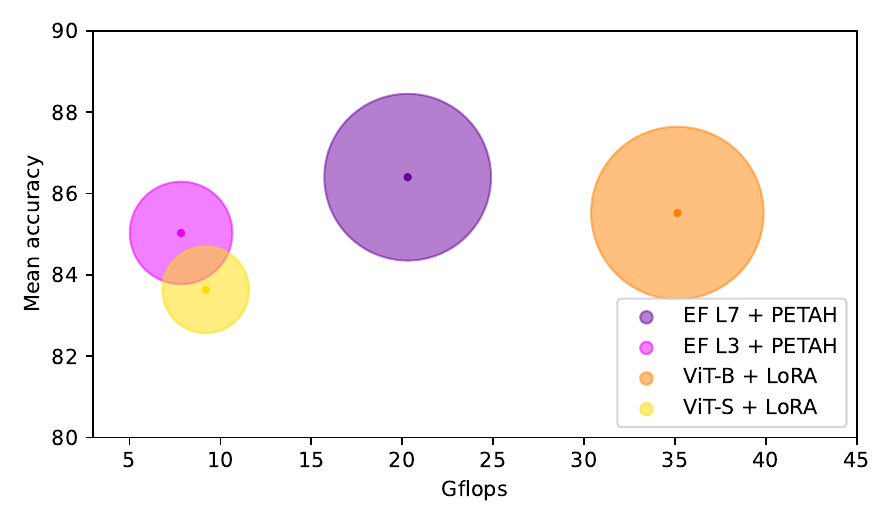}
     \end{subfigure}
     \vspace{-3mm}
    \caption{\textbf{Left:} Comparison of task-adaptation techniques, including PETAH for the EfficientFormer (EF) models. The number of adaptation parameters is plotted against mean accuracy on a fine-grained classification benchmark (refer to Table \ref{tab:classification}). Line colors indicate the adaptation technique, and markers represent model size. 
    \textbf{Right:} Performance of EF L7 and L3 models with PETAH-2 adaptation compared to ViT-B and ViT-S baselines using LoRA for attention layers. Mean accuracy is plotted against floating point operations (Gflops), with marker size proportional to backbone size. Note: PETAH-2 parameters for EF L7 (0.45M) are comparable to LoRA parameters for ViT-B (0.44M), and similarly for EF L3 and ViT-S (0.23M vs. 0.22M). 
    }
    \label{fig:fine_grained_main}
\end{figure}
\begin{table}
    \centering
    \caption{Task adaptation results for fine-grained classification using EfficientFormer models and ViT baselines. Mean accuracy is reported over three trials. Parameter counts exclude the linear head, whose size scales with the number of dataset classes ($K$ as $d \times K$). \petah-2 achieves the highest accuracy among all adaptation strategies for both EfficientFormer sizes, surpassing ViT models of comparable size. }
    \label{tab:classification}
    \setlength{\tabcolsep}{0.6em} 
    \begin{tabular}{c|l|cccccc|c|c}
    \hline
    & \textbf{Type} & \textbf{CUB} & \textbf{Cars} & \textbf{Pets} & \textbf{Aircraft} & \textbf{DTD} & \textbf{Food} & \textbf{Mean} & \textbf{\#Params}\\
    \hline
    \parbox[t]{2mm}{\multirow{7}{*}{\rotatebox[origin=c]{90}{EF L7}}}
    & Linear-probing & 88.01 & 70.48 & 93.52 & 52.98 & 75.60 & 86.65 & 77.88 & 0\\
    & ATTN FT & 88.40 & 88.14 & 93.76 & 68.58 & 76.54 & 89.12 & 84.36 & 15.7M\\
    & Full FT & \textbf{89.93} & 90.12 & {94.34} & 72.11 & 76.31 & {91.53} & 85.72 & 80.0M\\
    & LoRA ATTN $r=8$ & 88.47 & 88.57 & 93.96 & 69.20 & 76.83 & 89.12 & 84.36 & 0.26M\\
    & LoRA ATTN $r=16$ & 89.07 & 88.53 & 94.03 & 70.53 & 76.67 & 89.37 & 84.70 & 0.52M\\
    & LoRA ATTN+MLP $r=8$ & 89.65 & 87.16 & {94.34} & 69.20 & 77.27 & 89.31 & 84.49 & 0.75M\\
    & \petah-1 & 89.05 & 90.59 & 94.22 & 75.69 & 75.50 & 90.77 & 85.97 & 0.35M\\
    & \petah-2 & 89.07 & \textbf{91.20} & 94.19 & \textbf{75.96} & {77.32} & 90.66 & \textbf{86.40} & 0.45M\\
    \hline
    \parbox[t]{2mm}{\multirow{6}{*}{\rotatebox[origin=c]{90}{ViT-B}}}
    & Linear-probing & 88.98 & 75.00 & 94.11 & 54.63 & 76.47 & 88.68 & 79.65 & 0\\
    & ATTN FT &  {89.74} & 89.96 & \textbf{94.86} & 71.23 & \textbf{78.49} & 91.64 & {85.99} & 28.3M\\
    & Full FT & 89.48 & {90.08} & 94.69 & 69.70 & 78.17 & \textbf{91.67} & 85.63 & 85.7M\\
    & LoRA ATTN $r=8$ & 88.87 & 90.03  & 94.15 & {71.40} & 77.96 & 90.73 & 85.52 & 0.44M\\
    & Consolidator & 89.65 & 87.52 & 94.64 & 67.24 & 77.71 & 90.82 & 84.60 & 0.31M\\
    & SSF          & 89.48 & 87.82 & 94.79 & 67.33 & 77.64 & 89.50 & 84.43 & 0.21M\\
    \hline
    \multicolumn{10}{c}{}\\
    \hline
    \parbox[t]{2mm}{\multirow{5}{*}{\rotatebox[origin=c]{90}{EF L3}}}
    & Linear-probing & 85.15 & 66.94 & 91.59 & 50.10 & 74.01 & 84.57 & 75.39 & 0\\
    & ATTN FT & 87.92 & 84.52 & 93.48 & 66.80 & {75.35} & 87.27 & 82.56 & 5.25M\\
    & Full FT & \textbf{88.78} & 89.81 & 93.49 & 70.80 & 73.71 & \textbf{90.32} & 84.48 & 30.3M \\
    & LoRA ATTN  $r=8$ & 86.14 & 85.37 & 92.95 & 67.00 & 75.09 & 86.91 & 82.24 & 0.11M\\
    & \petah-1 & 87.91 & 89.50 & 93.68 & 73.11 & 74.38 & 89.02 & 84.60 & 0.17M \\
    & \petah-2 & 87.88 & \textbf{90.08} & {93.53} & \textbf{74.54} & 74.84 & 89.33 & \textbf{85.03} & 0.23M\\
    \hline
    \parbox[t]{2mm}{\multirow{6}{*}{\rotatebox[origin=c]{90}{ViT-S}}}
    & Linear-probing & 87.49 & 65.53 & 92.85 & 47.77 & 74.27 & 86.34 & 75.71 & 0\\
    & ATTN FT & {88.28} & 87.53 & \textbf{93.69} & {69.38} & \textbf{76.17} & 88.31 & {83.89} & 7.09M\\
    & Full FT & 87.69 & 87.93 & 93.62 & 66.85 & 75.30 & 83.94 &  83.56 & 21.6M\\
    & LoRA ATTN $=8$ & 86.86 & {88.09} & 93.12 & 68.45 & 75.03 & {88.60} & 83.36 & 0.22M\\
    & Consolidator & 88.01 & 83.85 & 93.26 & 62.49 & 75.48 & 88.37 & 81.91 & 0.10M\\
    & SSF          & 86.52 & 84.73 & 93.10 & 63.94 & 73.60 & 86.10 & 81.33 & 0.10M\\
    \hline
    \end{tabular}
\end{table}

\subsection{How to adapt a hybrid transformer}
We can now investigate the problem of adapting hybrid transformers. For standard language and vision transformers, it is common to only adapt the linear transformations in attention layers \cite{hu2021lora,he2023parameter}. However, in the case of the EfficientFormer and many other hybrid architectures that contain convolutional stages followed by attention-based stages \cite{dai2021coatnet, vasu2023fastvit, li2022rethinking}, such a procedure would only adapt the last part of the network and keep a large part of the signal path unchanged.
It is thus questionable whether such an adaptation is flexible enough to allow the model to adapt to various computer vision tasks or whether it is beneficial to also adapt the convolutional layers. For this experiment, we adapt the EF L7 backbone to three fine-grained classification datasets: FGVC-Aircraft \cite{maji2013fine}, Food101 \cite{bossard2014food} and the Describable Textures Dataset (DTD) \cite{cimpoi14describing}.

For each task, we add a linear head on top of the frozen backbone and use PEFT to adapt specific modules. We compare various adaptation strategies:

\textbf{Linear Probing:} Only the linear head is trained; the backbone remains frozen.

\textbf{LoRA on Attention Layers:} Adapts only the attention weights using LoRA with varying ranks.

\textbf{LoRA on Attention and MLP Layers:} Adapts both attention and MLP layers.

\textbf{LoRA on Attention and Convolution layers:} Adapts convolutional layers using convolutional LoRA (\ref{eq:conv_lora}) with low ranks ($r_c = 1$ or $2$) and attention layers using standard LoRA. MLP layers stay frozen.

Note that when we adapt the attention weights, we always refer to the $W_Q,W_K,W_V$ weight matrices and the surrounding projection layers in the EfficientFormer. We tune the learning rate and weight decay for each method using a grid search on a separate validation set and report the test accuracy for the best-performing configuration. Since Food101 does not have an extra validation split, we create one from the train set by separating 50 examples for each class. Results can be found in Table \ref{tab:classification_lora_layers}.

We notice several interesting findings that differ from the PEFT literature for pure transformer architectures. First, we note that for the EfficientFormer L7, adapting the MLP weights does increase downstream task performance, which was found to not be the case for ViTs \cite{he2023parameter}. However, while such adaptation can be beneficial to performance, due to the high dimensionality of the MLP matrices, this approach significantly increases the total parameter count by a factor of 3 over the attention-only adaptation from 260K to 750K. On the other hand, we show that adapting the convolutional layers with a low rank of $r_c = 1$ or $2$ can significantly outperform adaptation methods that only change the weights of the attention or MLP layers. In total, adding LoRA adaptation to convolutional layers requires about 100K-200K extra parameters depending on the rank of the convolutional adaptation $r_c$ and creates the best-performing method in mean accuracy over all tasks. It even outperforms adaptations with more parameters, such as high-rank LoRA for the attention module or LoRA for the attention and MLP module, implying that adapting the entire forward pass of the network is more beneficial than focusing purely on the last stage containing the transformer blocks. 

While adapting the MLP modules in the Meta3D blocks in stage 4 and higher rank adaptations are beneficial, we quickly reach diminishing returns in terms of the accuracy parameter trade-off. For LoRA ATTN, doubling the rank from 8 to 16 causes an improvement in mean accuracy from 78.38\% to 78.75\% at the cost of doubling the parameter count, and for LoRA ATTN + MLP, rank 16 adaptation reaches the highest mean accuracy of 78.78\% at the cost of requiring 1.5M parameters per task. Attention adaptation using LoRA combined with our convolutional LoRA for the Meta4D blocks outperforms all adaptations that only modify the Meta3D blocks, even the ones with up to 3 times as many parameters, implying that for hybrid transformers, attention tuning is \textit{not} all you need. 

\subsection{PETAH: Parameter Efficient Task Adaptation for Hybrid transformers}
We now introduce \petah, our PEFT framework for hybrid transformers. It uses standard LoRA adaptation of attention layers combined with low-rank convolution adaptation for all convolutional layers, as introduced earlier. Particularly, we define PETAH-$r_c$ as adapting linear layers inside attention module with LoRA of rank $r=8$ and all convolutions with convolutional LoRA of rank $r_c$. Any other fully connected layers outside of attention layer are not modified (see Figure \ref{fig:teaser}).
As seen in previous section, this setup strikes a good balance between performance and parameter efficiency. 
Our method exploits the fact that, unlike MLP modules which usually use a LoRA rank between $4$ and $16$, convolutional layers can be effectively adapted with a low rank ($r_c = 1$ or $2$). This remarkably boosts performance with a relatively small number of additional parameters, as it allows us to adapt the entire forward pass of the model.

\noindent{\textbf{Parameter Complexity:}} Regarding additional parameters, let $d_k$ be projection dimension of $W_Q$ and $W_K$ and $d_v$ be value dimension of $W_V$. Further, let $d$ be equal to input dimension of attention layer. In \ref{eq:lora}, for adapting $W_Q,W_K$ and $W_V$, all matrices $A$ will be of dimension $r \times d$ and matrices $B$ of sizes $d_k \times r$ and $d_v \times r$ respectively. Thus, to adapt one attention layer with $h$ heads using LoRA, we require $hr (2d_k + d_v + 3d)$ additional parameters. As we also adapt linear projection in MHSA layer, we require additional $r(d + hd_v)$  parameters. For \petah-$r_c$ on a convolutional layer of kernel size $k$, $p$ input, and $q$ output channels, our method adds $r_c qk^2 + r_c p$ learnable parameters.  

\begin{table}
    \centering
    \caption{EF L7 task-adaptation for object detection and instance segmentation using Cascade R-CNN on COCO and semantic segmentation on ADE20K using Semantic FPN.}
    \label{tab:detection}
    \begin{tabular}{c|l|cc|cc|c}
    \hline
        & Backbone & \multicolumn{2}{c|}{Object Detection} & \multicolumn{2}{c|}{Instance Segmentation} & \multicolumn{1}{c}{Semantic}\\
         & Adaptation & $\mathbf{AP}^\textrm{box}$ & $\mathbf{AR}^\textrm{box}$ & 
         $\mathbf{AP}^\textrm{mask}$ & $\mathbf{AR}^\textrm{mask}$ & \textbf{mIoU} \\
         \hline
         \parbox[t]{2mm}{\multirow{5}{*}{\rotatebox[origin=c]{90}{EF L7}}} & 
         Frozen  & 0.32 & 0.46 & 0.31 & 0.45 & 31.2\\
         & ATTN FT & 0.38 & 0.50 & 0.36 & 0.48 & 43.0 \\ 
         & Full FT & \textbf{0.41} & 0.51 & \textbf{0.38} & 0.47 & \textbf{48.3}\\
         & LoRA ATTN $r=8$ & 0.36 & 0.50 & 0.35 & 0.47 &	40.3\\
         & \petah-1  & 0.39 & 0.51 & 0.37 & \textbf{0.49} &	44.2\\
         & \petah-2 & 0.39 & \textbf{0.52} & 0.37 & \textbf{0.49} & 45.0\\
         \hline

    \end{tabular}
\end{table}
\section{Experiments}
\label{sec:5_Experiments}

In the following section, we will evaluate \petah on several vision benchmarks, including classification, object detection, and semantic segmentation, and compare it to other task-adaptation techniques. We will also compare different model sizes and, in particular, evaluate the performance of \petah for hybrid models compared to ViT adaptation.

\subsection{Classification}
For fine-grained classification, we again use Aircraft, DTD, and Food101 datasets and additionally evaluate on CUB200 \cite{wah2011caltech}, Oxford-IIIT Pets \cite{parkhi2012cats} and Stanford Cars \cite{krause2013cars}.
Since we found that hyperparameters like learning rate of the linear head, learning rate of the adapter, weight decay, and number of epochs can strongly influence the results and do not necessarily transfer between different adaptation strategies, we use the validation split to find the best-performing parameters for each method. 
Please see \ref{tab:app_hparams} in the supplementary section for a detailed listing of all hyperparameter configurations for all methods.
Since CUB, Pets, and Cars do not have a separate validation split and are relatively small, we reuse the best-performing parameters from Aircraft and DTD since they are most similar in size.
We use standard data augmentation, including random crops and horizontal flip. After hyperparameter selection, we run each experiment with 3 different, fixed random seeds and report the average performance. 

For the EfficientFormer models, we compare linear-probing, full fine-tuning, attention tuning \cite{touvron2022three}, LoRA applied to attention, as well as our \petah with convolutional rank $r_c$ equal to 1 (\petah-1) and 2 (\petah-2). Unless otherwise specified, all LoRA adaptations for fully connected layers, including non-convolutional adaptation in \petah, use rank 8. For EF L7, we include LoRA with rank 16 and LoRA for attention and MLP layers as baselines. 

For ViTs, we use linear probing, full fine-tuning, attention tuning, and LoRA in a common configuration of rank 8 applied to attention layers. We compare to the recent Consolidator \cite{hao2023consolidator} and Scaling \& Shifting Your Features (SSF) \cite{lian2022scaling} which are both PEFT methods designed for the vision domain, specifically, ViT architecture. Consolidator uses grouped connections across features to construct tunable parts and includes stochastic depth in the adaptation path as regularization. As in the original paper, Consolidator is applied to all linear layers in a ViT, including the MLP layers. SSF operates by scaling and shifting the features within the network, and it is applied after every operation in the ViT.

Results can be found in Figure \ref{fig:fine_grained_main} and Table \ref{tab:classification}. We find that for all EfficientFormer models, PETAH clearly outperforms all other adaptation methods in terms of mean accuracy, including full model fine-tuning, which requires more than 150 times as many parameters. Importantly, the combination of PETAH with the EF7 model outperforms all ViT-B-based approaches when comparing backbones trained with the same pre-training setup. As mentioned in Section \ref{sec:hybrid_architectures_and_pre_training}, these two models have a comparable number of parameters and pre-training accuracy, however, the EF L7 has a smaller FLOP count and on-device latency (6.9ms to 18.2ms for ViT-B) and is thus better suited for mobile applications. 

We are the first to demonstrate that hybrid models can be adapted to new downstream tasks more effectively than ViTs.  This is only possible with our \petah approach specifically tailored for hybrid architecture as all other adaptation strategies on EF generally perform worse than LoRA-based approaches for ViTs, for example, EF L7 with conventional PEFT is unable to achieve a mean accuracy higher than 85\%, whereas a ViT-B with LoRA achieves 85.52\%. Both of our \petah variants surpass them with 85.97\% mean accuracy for $r_c = 1$ and 86.4\% for $r_c = 2$ at a comparable number of additional parameters ($\sim0.45$M for EF L7 with \petah-2 and ViT-B with LoRA). We also note that in terms of full fine-tuning, EF L7 (85.72\% mean accuracy) shows a very similar performance to a fully fine-tuned ViT-B (85.63\% mean accuracy), demonstrating that \petah improvements come from a stronger task-adaptation performance and are not caused by a stronger base model. 

Compared to Consolidator \cite{hao2023consolidator} and SSF \cite{lian2022scaling}, two methods specifically developed for visual adaptation for vision transformers, we can see that our \petah-adapted EfficientFormer clearly outperforms baseline. In particular, for EF L7 with \petah-1 and 2, we achieve a mean accuracy of 85.97\% and 86.40\% compared to 84.60\% for Consolidator and 84.43\% for SSF. 
To validate our Consolidator results, we compare our results on our own ViT-B to their reported accuracies on same architecture with different pre-training and note that we achieve better results on all shared datasets:

\begin{center}
    \begin{tabular}{c|ccc|c}
\hline
  \textbf{Consolidator} & CUB & DTD & Pets & Mean \\ 
  \hline
   Our results &  \textbf{89.7} &  \textbf{77.7} & \textbf{94.6} & \textbf{87.3} \\
   Reported  results &  87.0 & 74.5 & 92.3 & 84.6\\
\hline
\end{tabular}

\end{center}

For the smaller models, the EF L3 with \petah clearly outperforms the ViT-S baseline (EF L3 with \petah-2 mean acc.: 85.05\% vs 83.89\% for the best-performing ViT-S adaptation) while having 4 times smaller NPU latency and the additional adaptation parameters of \petah-2 for the EF L3 are comparable to LoRA for a ViT-S. This also includes Consolidator with a mean accuracy of $81.91\%$ and SSF with a mean accuracy of 81.33\%. In general, while both require fewer parameters than LoRA, we did not find them to yield a higher mean accuracy for either the ViT-B or ViT-S.

\subsection{Extension to sparse backbones}
While the EfficientFormer (EF) architecture is efficient in terms of on-device latency~\cite{li2022efficientformer}, the L7 variant with 80 million parameters is still too large for many resource-limited applications. To address this, we experiment with combining PEFT and sparsity. By integrating our \petah framework with pruning, we achieve a sparse and efficient backbone that can be easily adapted to new tasks. It is worth noting that starting from large models and applying aggressive pruning is an established technique~\cite{zhu2017prune} that often yields more performant models compared to less aggressive pruning of smaller models.
For pre-training the sparse models, we use the same general training setup outlined in Section \ref{sec:hybrid_architectures_and_pre_training}. We train both an EfficientFormer L7 and a ViT-B from scratch using Spartan~\cite{tai2022spartan} with a 90\% sparsity ratio for both models. The resulting models perform as follows compared to their dense counterparts:

\begin{center}
\begin{tabular}{c|ccccc}
        \hline
        \textbf{Model} & \textbf{Sparsity} & \textbf{IN21K acc.} & \textbf{\#params}\\
        \hline
        EF L7 &  - & 50.43 & 80M\\ 
        EF L7 & 0.9 & 47.11 & 7.9M\\ 
        \hline
        ViT-B & - & 49.99 & 85M\\ 
        ViT-B &  0.9 & 46.98 & 8.1M\\ 
        \hline
\end{tabular}
\end{center}
Details of pruning strategy is discussed in appendix section.

\subsection{Object detection and semantic segmentation} 

While other works on PEFT mostly focus on  LLMs, classification~\cite{he2023parameter}, or generation tasks~\cite{yeh2023navigating}, our goal is to demonstrate that hybrid backbones with \petah can handle a wide range of computer vision applications, including dense prediction tasks. Object detection and segmentation rely on different feature representations compared to classification, making it important to assess whether \petah is flexible enough to adapt to these tasks.

For object detection and instance segmentation, we follow a similar setup to~\cite{li2022efficientformer, li2022next}, using Cascade R-CNN~\cite{cai2018cascade} with Feature Pyramid Network (FPN) \cite{lin2017feature} on COCO for $640 \times 480$ resolution.
For semantic segmentation, we use the challenging ADE20K dataset~\cite{zhou2017scene}, which consists of 20K training images covering 150 classes. We employ our pretrained EfficientFormer L7 backbone and attach a Semantic FPN~\cite{kirillov2019panoptic} on top, combining with different PEFT methods.

One major advantage of hybrid models over ViTs is their hierarchical feature representation, which allows for multi-scale representations utilized by many classic computer vision algorithms~\cite{lin2017feature, cheng2021mask2former}. Since ViTs do not naturally produce hierarchical feature maps and typically require large adapters to be effective in dense prediction tasks~\cite{chen2022vision, li2022exploring}, we exclude them from this experiment.

Results for the EfficientFormer L7 are presented in Table \ref{tab:detection}. While full fine-tuning remains the best-performing method, our \petah approach clearly outperforms LoRA and attention fine-tuning, coming in as a close second to standard full fine-tuning—which requires more than 170 times as many parameters. Specifically, on COCO, \petah-2 achieves the highest mean Average Recall (mAR) for both bounding boxes (0.52) and instance masks (0.49). It also attains the second-highest mean Average Precision (mAP) with 0.39 for detection and 0.37 for instance segmentation, only slightly underperforming compared to the parameter-heavy full fine-tuning (detection mAP: 0.41, instance segmentation mAP: 0.38).

For semantic segmentation on ADE20K, we observe similar results. \petah-2 achieves a mean Intersection over Union (mIoU) of 45.0, significantly outperforming LoRA adaptation (40.3 mIoU) and attention fine-tuning (43.0 mIoU). Overall, \petah outperforms all other adaptation methods except full fine-tuning on dense vision tasks, demonstrating its effectiveness and versatility.

\section{Conclusion and limitations}
\label{sec:Conclusion and Limitations}

In this work, we introduced \petah, a PEFT framework for hybrid transformers that modifies not only the attention layers but also the convolutional layers, clearly outperforming baseline PEFT approaches. The resulting adapted models can surpass ViTs of comparable size while being more computationally efficient. Additionally, we demonstrated that for sparse hybrid backbones, \petah adaptation outperforms even full fine-tuning and can recover part of the performance loss caused by pruning. 
Due to their hierarchical feature maps, hybrid backbones can be easily adapted to various vision tasks. Our results on object detection and semantic segmentation tasks confirm that \petah effectively extends to dense prediction tasks without additional inference-time overhead.

Regarding limitations, we note that we restricted most of our analysis to EfficientFormer backbone since the backbone and resulting PEFT adaptations are comparable in terms of number of parameters. While it is possible to manually extend \petah to other hybrid backbones and employ more efficient PEFT factorizations~\cite{edalati2022krona, yeh2023navigating}, ideally one should combine convolutional adaptation with a random-search-based approach like GLORA~\cite{chavan2023one} to automatically find an ideal variant without need for manual configuration.

\clearpage


\bibliographystyle{unsrt}  
\bibliography{main}  


\end{document}